\documentclass[letterpaper, 10 pt, journal, twoside]{IEEEtran}

\usepackage{graphics} 
\usepackage{epsfig} 
\usepackage{amsmath} 
\usepackage{amssymb}  
\usepackage{algorithm}
\usepackage{algorithmic}
\usepackage{mathrsfs}
\usepackage{mathtools}
\usepackage{multirow}
\usepackage{comment}
\usepackage{color}
\usepackage{cite}
\usepackage{url}
\usepackage{soul}
\usepackage{listings}
\usepackage[utf8]{inputenc}
\usepackage[T1]{fontenc}
\usepackage[english]{babel}
\usepackage{xcolor}
\usepackage[normalem]{ulem}

\usepackage{xpatch} 

\newtheorem{theorem}{Theorem}

\newtheorem{rem}{Remark}
\newtheorem{lemma}{Lemma}
\newtheorem{assumption}{Assumption}

\begin{document}
\newcommand{\vect}[1]{\boldsymbol{\mathbf{#1}}}
\newcommand{\margin}[1]{\marginpar{\color{red}\tiny\ttfamily#1}}
\newcommand{\solmaz}[1]{{\color{red}#1}}
\newcommand{\blue}[1]{{\color{blue}#1}}
\makeatletter
\renewcommand*{\@opargbegintheorem}[3]{\trivlist
      \item[\hskip \labelsep{ #1\ #2}] (#3):\ \itshape}
\makeatother
 \newcommand{\boxend}{\hfill \ensuremath{\Box}}
\title{Measurement Scheduling for Cooperative Localization in Resource-Constrained Conditions}

\author{Qi Yan$^{1}$, Li Jiang$^{2}$ and Solmaz S. Kia$^{3}$, \emph{Senior Member, IEEE}
\thanks{Manuscript received: September 10, 2019; Revised December 9, 2019; Accepted January 7, 2020.
This paper was recommended for publication by Editor S. Behnke upon evaluation of the Associate Editor and Reviewers' comments.
The work of S. S. Kia is supported by the U.S. Dept. of Commerce, National Institute of Standards and Technology award 70NANB17H192.
} 
\thanks{$^{1}$Qi Yan is with the Institute of Mechanical Engineering, École Polytechnique Fédérale de Lausanne EPFL, Lausanne 1015, Switzerland. {\tt\small qi.yan@epfl.ch}}
\thanks{$^{2}$Li Jiang is with the Department of Computer Science and Engineering, Shanghai Jiao Tong University, Shanghai 200240, China.           {\tt\small jiangli@cs.sjtu.edu.cn}}%
\thanks{$^{3}$Solmaz S. Kia is with the Department of Mechanical and Aerospace Engineering, University of California Irvine, Irvine, CA 92697, USA,
	{\tt\small solmaz@uci.edu}}%
\thanks{Digital Object Identifier (DOI): see top of this page.}
}

\markboth{IEEE Robotics and Automation Letters. Preprint Version. Accepted January, 2020}
{Yan \MakeLowercase{\textit{et al.}}: Measurement Scheduling for Cooperative Localization in Resource-Constrained Conditions}

\maketitle

\begin{abstract}
This paper studies the measurement scheduling problem for a group of $N$ mobile robots moving on a flat surface that are preforming cooperative localization (CL). We consider a scenario in which due to the limited on-board resources such as battery life and communication bandwidth only a given number of relative measurements per robot are allowed at observation and update stage. Optimal selection of which teammates a robot should take a relative measurement from such that the updated joint localization uncertainty of the team is minimized is an NP-hard problem.
In this paper, we propose a suboptimal greedy approach that allows each robot to choose its landmark robots locally in polynomial time. Our method, unlike the known results in the literature, does not assume full-observability of CL algorithm. Moreover, it does not require inter-robot communication at scheduling stage.  That is, there is no need for the robots to collaborate to carry out the landmark robot selections. 
We discuss the application of our method in the context of an state-of-the-art decentralized CL algorithm and demonstrate its effectiveness through numerical simulations. 
Even though our solution does not come with rigorous performance guarantees, its low computational cost along with no communication requirement makes it an appealing solution for operations with resource constrained robots.
\end{abstract}

\section{Introduction}
\label{sec::intro}
\IEEEPARstart{W}{e} consider the problem of relative measurement scheduling in  cooperative localization (CL) for a team of mobile robots. In cooperative localization mobile robots improve their localization accuracy by jointly processing the  relative measurements they take with respect to each other~\cite{roumeliotis2002distributed}. 
Cooperative localization is of interest in operations where access to global positioning signal (GPS) and external landmarks for positioning aid (to conduct SLAM) are challenging, e.g., in underwater operations~\cite{SEW-JMW-LLW-RME:13,bahr2009consistent} or in indoor localization for firefighters~\cite{JN-JR-PH-IS-MO-KVSH:14,zhu2018loosely}. In the past two decades CL has been studied extensively in the literature with the the focus being mainly on the design of decentralized CL algorithms, e.g.,~\cite{roumeliotis2002distributed,bahr2009consistent,carrillo-arce2013decentralized,SSK-SF-SM:16,kia2018serverassisted,luft2018recursive,ZJ-SSK:19}. To process every inter-agent relative measurement, the robots need to communicate with each other. In decentralized algorithm designs, the focus is to remove the need for all-to-all communication to reduce the communication load. The communication cost however still can be high if the robots take measurements from every robot that is in their measurement~zone.  

The problem of computation/communication cost reduction via smart management of inter-robot relative measurements has been studied in~\cite{chang2018optimal,mourikis2006optimal,caglioti2006cooperative,zhang2018multi,singh2017supermodulara}.
One of the approaches used is to determine the optimal measurement frequency given resource constraints~\cite{chang2018optimal,mourikis2006optimal}. The methods used in~\cite{chang2018optimal} and~\cite{mourikis2006optimal} assume a fixed sensing topology throughout the operation and  use covariance upper bound analysis to determine the relative measurement frequency in the team. These methods use the steady-state covariance upper bound obtained by the Discrete-time Algebraic Riccati Equation (DARE) and the corresponding Continuous Algebraic Riccati Equation (CARE). 
The convergence of the Riccati recursion requires that the overall system is observable \cite{chang2018optimal,mourikis2006optimal}, which means there must be at least one robot accessing absolute positioning information such as GPS signals or known landmarks~\cite{mourikis2006performance}.
The observability requirement is however a very hard constraint and cannot be satisfied in many conditions such as in uncharted indoor environments or underwater operations. The method used in~\cite{chang2018optimal} relies on exhaustive search, which may not be a practical choice for real-time implementation.

Another approach for cost management in CL is via measurement scheduling, in which robots get restricted to take only certain number of relative measurements but they have to choose their landmark robots\footnote{A landmark robot is a robot that another robot takes measurement from.} in a way that the positioning uncertainty is  minimized. Some results on measurement scheduling can be found in~\cite{caglioti2006cooperative,zhang2018multi,singh2017supermodulara}. The uncertainty measure that is minimized in these work is mainly the logarithm of the determinant (logdet) of the joint  covariance matrix of the team, which is indeed a measure of the volume of the uncertainty ellipsoid~\cite{tzoumas2016nearoptimal}. It is known in the literature that the optimal measurement scheduling for CL is an NP-hard problem~\cite{singh2017supermodulara,zhang2017sensor}. Therefore, the main effort is on proposing suboptimal solutions with reasonable computational complexity. The studies in~\cite{caglioti2006cooperative,zhang2018multi} use some form of greedy algorithms to carry out the landmark robot selections. Tzoumas et al.~\cite{tzoumas2017scheduling,tzoumas2016nearoptimal} and Zhang et al.~\cite{zhang2017sensor} analyze the supermodularity of the logdet of the joint covariance matrix to propose suboptimal greedy measurement scheduling solutions with known optimality gap. They also  investigate the time complexity of their proposed greedy algorithm for a Kalman filter based CL algorithm. In an alternative approach, Singh et al. \cite{singh2017supermodulara} employ the trace of the joint covariance as the objective function and propose a suboptimal greedy solution for the measurement scheduling of multi-robot CL.
The above methods whether using logdet or trace of the joint covariance as objectives function however, suffer from high inter-robot communication cost in scheduling process.
Moreover, they all require real-time robot locations to compute observation matrices for scheduling purpose. 
Sensor scheduling is of interest in other localization algorithms, as well. Interested reader can see ~\cite{cieslewski2018data,tian2019resource} for some recent~studies.

In this letter, we present a novel method for relative measurement scheduling in a CL algorithm that allows the robots moving on a flat surface to decide locally what other team members to take relative measurements from so that a high accuracy is achieved when they are constrained to take limited number of measurements. Our special focus is on design of a measurement scheduling algorithm that does not require inter-robot communication.  That is, there is no need for the robots to collaborate to carry out the landmark robot selections. We also seek a solution that does not assume full-observability, and can be carried with reasonable computational complexity in real-time.
Our proposed solution is a greedy landmark selection heuristic that works based on minimizing an upper-bound on the total uncertainty of the team.
We use the logdet of the joint covariance matrix as our uncertainty measure to reduce while conducting measurement scheduling. 
We show through numerical examples that, even though our proposed solution does not have rigorous performance guarantees, its application results in a localization performance comparable to that achieved by the landmark selection algorithm of~\cite{tzoumas2017scheduling} that comes with a known optimality gap. 
We provide a computational complexity analysis and show that our algorithm has considerable lower computational cost than that of~\cite{tzoumas2017scheduling}. This low computational cost along with no communication requirement makes our algorithm an attractive landmark selection solution for operations with resource constrained~robots.

\section{Problem formulation and objective statement}
\label{sec::prob}
In this section, we review the joint Extended Kalman Filter (EKF) CL for position estimation on which we place our work. Then, we present our objective statement. 

\subsection{Problem formulation}
Consider a group of $N$ robots moving on a flat terrain. The equation of motion of each robot $i\in\mathcal{V}=\{1,\cdots,N\}$ is described by  %
\begin{equation}
\begin{split}
\boldsymbol{x}^i(k+1) &= \boldsymbol{f}^i(\boldsymbol{x}^i(k),v^i(k),\phi^i(k))	
\\&= \boldsymbol{x}^i(k)+\begin{bmatrix}
 \delta t\, v^i(k)  \cos(\phi^i(k)) \\ 
 \delta t\, v^i(k)  \sin(\phi^i(k))
\end{bmatrix},
\end{split}
\label{equ::EKF_sys_motion}
\end{equation}
where state $\boldsymbol{x}^i(k)=[x^i(k),y^i(k)]^\top$ is the absolute position of the robot with respect to a global map.
Here, $v^i$ is the linear motion velocity, $\phi^i$ is the robot's orientation and $\delta t>0$ is the stepsize. 
Each robot $i\in\mathcal{V}$ uses a wheel encoder to measure its linear velocity $v_m^i=v^i+\eta_v^i$, and a compass to compute its   absolute orientation $\phi_m^i=\phi^i+\eta_\phi^i$
to propagate its states according to
\begin{equation}
\begin{split}
\hat{\boldsymbol{x}}^{i-}(k+1) &= \boldsymbol{f}^i(\hat{\boldsymbol{x}}^{i+}(k),v^i_m(k))
\\&= \hat{\boldsymbol{x}}^{i+}(k)+\begin{bmatrix}
\delta t\,  v^i_m(k)  \cos({\phi}_m^i(k)) \\ 
 \delta t\,v^i_m(k)   \sin({\phi}_m^i(k))
\end{bmatrix}.
\end{split}
\label{equ::EKF_prop_x}
\end{equation}
 Here, $\eta_v^i\sim N(0,\sigma_{v^i})$ and $\eta_\phi^i\sim N(0,\sigma_{\phi^i})$ 
are the white zero-mean Gaussian noises contaminating, respectively, the linear velocity and orientation measurements.
Hereafter, the superscript $^+$ and $^-$ stand for a posterior (updated) and a priori (propagated) estimation, respectively. Let $\tilde{\boldsymbol{x}}^{i-}(k) = \boldsymbol{x}^{i}(k) - \hat{\boldsymbol{x}}^{i-}(k)$ and $\tilde{\boldsymbol{x}}^{i+}(k) = \boldsymbol{x}^{i}(k) - \hat{\boldsymbol{x}}^{i+}(k)$ be, respectively, the propagated and updated state errors. Using the motion dynamics~\eqref{equ::EKF_sys_motion}, we obtain
\begin{equation*}
\begin{split}
&\tilde{\boldsymbol{x}}^{i-}(k+1) = \tilde{\boldsymbol{x}}^{i+}(k) \\&\qquad + \delta t \begin{bmatrix}
{\cos}({\phi}_m^i(k)) & -v_m^i(k)\,{\sin}({\phi}_m^i(k))\\
{\sin}({\phi}_m^i(k)) & v_m^i(k)\,{\sin}({\phi}_m^i(k))
\end{bmatrix}
\begin{bmatrix}
\eta_v^i(k) \\ \eta_{\phi}^i(k)
\end{bmatrix}.
\end{split}
\label{equ::EKF_prop_x_tilde}
\end{equation*}
Then, the propagated error covariance using the linearized model of the motion dynamics~\eqref{equ::EKF_sys_motion} is 
\begin{subequations}\label{equ::EKF_prop_P}
\begin{align}
\boldsymbol{P}^{i-}(k+1) &= \boldsymbol{P}^{i+}(k) + \boldsymbol{Q}^i(k),\label{equ::EKF_prop_Pi}\\
\boldsymbol{P}^{-}_{ij}(k+1) &= \boldsymbol{P}^{+}_{ij}(k),~~\qquad j\in\{1,\cdots,N\}\backslash\{i\},\label{equ::EKF_prop_Pij}
\end{align}
\end{subequations}
where $i\in\mathcal{V}$ and the incremental term $\boldsymbol{Q}^i(k)$ is
\begin{equation}
\boldsymbol{Q}^i(k) = (\delta t)^2 \boldsymbol{C} ({\phi}_m^i(k)) \begin{bmatrix}
\sigma_{v^i}^2 & 0\\
0 & (v^i_m)^2 \sigma_{{\phi}^i}^2
\end{bmatrix}
\boldsymbol{C}^\top ({\phi}_m^i(k)).
\label{equ::EKF_prop_P_Q}
\end{equation}
Here, $\boldsymbol{C}(\phi)$ is the rotational matrix with respect to orientation $\phi$. Next, let the relative measurement taken by robot $a$ from  robot $b$ at timestep $k$, denoted by $a\!\overset{k}{\longrightarrow}\! b$, be the relative~position
\begin{align}
\label{equ::EKF_sys_mea} \boldsymbol{z}_{ab}(k) &= \boldsymbol{h}_{ab}(\boldsymbol{x}^a(k),\boldsymbol{x}^b(k)) + \boldsymbol{\nu}^{ab}(k)
\\&= \boldsymbol{C}^\top(\phi^a(k)) \left(\begin{bmatrix}
x^b(k)\\y^b(k)
\end{bmatrix} - \begin{bmatrix}
x^a(k)\\y^a	(k)
\end{bmatrix} \right)+ \boldsymbol{\nu}^{ab}(k) \nonumber
\end{align}
where $\boldsymbol{\nu}^{ab}$ is the white-Gaussian measurement noise. The relative position measurement is obtained from inter-robot ranging and bearing  sensors on robot $a$, e.g., via a Kinect camera and augmented reality (AR) tags~\cite{niekum2013ar_track_alvar}. In other words,
\begin{equation*}
	\boldsymbol{z}_{ab}(k) = \rho^{ab}(k)\begin{bmatrix}
	{\rm cos}(\phi^{ab}(k)) \\ {\rm sin}(\phi^{ab}(k))
	\end{bmatrix} + \boldsymbol{\nu}^{ab}(k)
	\label{equ::EKF_sys_mea_r+b}
\end{equation*}
	where $\rho^{ab}$ and $\phi^{ab}$ are, respectively, the true relative range and relative bearing between robots $a$ and $b$. Let $\rho_m^{ab} = \rho^{ab} + \eta_{\rho^a}$ and $\phi^{ab}_m = \phi^{ab} + \eta_{\theta^a}$ be, respectively, the measured relative range and relative bearing contaminated by measurement noises 
$\eta_{\rho^a} \sim N(0,\sigma_{\rho^a})$ and $\eta_{\theta^a} \sim N(0,\sigma_{\theta^a})$. 
By linearizing \eqref{equ::EKF_sys_mea}, the measurement innovation $\tilde{\boldsymbol{z}}_{ab}(k) = \boldsymbol{z}_{ab}(k) - \boldsymbol{h}_{ab}(\hat{\boldsymbol{x}}^{a-}(k),\hat{\boldsymbol{x}}^{b-}(k)) $ can be obtained as
\begin{equation*}
\begin{split}
& \tilde{\boldsymbol{z}}_{ab}(k) = \boldsymbol{H}_{ab,a}(k) \tilde{\boldsymbol{x}}^{a-}(k) + \boldsymbol{H}_{ab,b}(k) \tilde{\boldsymbol{x}}^{b-}(k) 
\\&+ \boldsymbol{C}^\top(\phi^a_m(k))
\boldsymbol{J}
( \hat{\boldsymbol{x}}^{b-}(k) - \hat{\boldsymbol{x}}^{a-}(k)) \eta_\phi^a(k)
 + \boldsymbol{\nu}^{ab}(k)
\end{split}
\end{equation*}
with
$\boldsymbol{H}_{ab,a}(k) = -\boldsymbol{C}^\top(\phi^a_m(k))$,
$\boldsymbol{H}_{ab,b}(k) = \boldsymbol{C}^\top(\phi^a_m(k))$
and $ \boldsymbol{J} = \begin{bmatrix}
0 & 1 \\ -1 & 0 \end{bmatrix}$.
The covariance of innovation $\tilde{\boldsymbol{z}}_{ab}(k)$ can be written as~\cite{mourikis2006performance}
\begin{equation*}
\begin{split}
\boldsymbol{S}_{ab}(k) =\,&\,
\boldsymbol{H}_{ab,a}(k)\boldsymbol{P}^{a-}(k)\boldsymbol{H}_{ab,a}(k)^\top+
\\ & \boldsymbol{H}_{ab,a}(k)\boldsymbol{P}_{ab}^{-}(k)\boldsymbol{H}_{ab,b}(k)^\top+
\\ &\boldsymbol{H}_{ab,b}(k)\boldsymbol{P}^{b-}(k)\boldsymbol{H}_{ab,b}(k)^\top+
\\ & \boldsymbol{H}_{ab,b}(k)\boldsymbol{P}_{ba}^{-}(k)\boldsymbol{H}_{ab,a}(k)^\top+
\\ &  \boldsymbol{R}_{\phi^{ab}_m}(k) + \boldsymbol{R}_{z_{ab}}(k), 
\end{split}
\end{equation*}
where
\begin{subequations}
\begin{equation}
\begin{split}
\boldsymbol{R}_{\phi^{ab}_m}(k) &= \sigma_{\phi^a}^2\boldsymbol{C}^\top(\phi^{a}_m(k))
\boldsymbol{J}
( \hat{\boldsymbol{x}}^{b-}(k) - \hat{\boldsymbol{x}}^{a-}(k))
\\&\times
( \hat{\boldsymbol{x}}^{b-}(k) - \hat{\boldsymbol{x}}^{a-}(k))^\top
\boldsymbol{J}^\top
\boldsymbol{C}(\phi^{a}_m(k)),
\end{split}
\end{equation}
\begin{equation}
\begin{split}
&\boldsymbol{R}_{z_{ab}}(k)=\mathsf{E}[\boldsymbol{\nu}^{ab}(k)\boldsymbol{\nu}^{ab}(k)^\top]
\\&= \boldsymbol{C}(\phi^{ab}_m(k))
\begin{bmatrix}
\sigma_{\rho^a}^2 & 0 \\
0 & (\rho_m^{ab}(k)\sigma_{\theta^a})^2
\end{bmatrix}
\boldsymbol{C}^\top(\phi^{ab}_m(k)).
\end{split}
\end{equation}
\label{equ::EKF_update_S_R}
\end{subequations}
We note that $\boldsymbol{R}_{\phi^{ab}_m}$ and $\boldsymbol{R}_{z_{ab}}$ are due to local absolute orientation measurement error and relative position measurement noise, respectively. Given the measurement innovation and its corresponding covariance matrix $\boldsymbol{S}_{ab}(k)$, following the standard EKF framework, we now obtain the state and covariance update as
\begin{subequations}\label{eq::Joint_update}
\begin{equation}
\hat{\boldsymbol{x}}^{i+}(k+1) = \hat{\boldsymbol{x}}^{i-}(k+1) + \boldsymbol{K}_i(k+1)\tilde{\boldsymbol{z}}_{ab}(k+1),
\label{equ::EKF_update_x}
\end{equation}
\begin{equation}
\begin{split}
\boldsymbol{P}^{i+}(k+1) &= \boldsymbol{P}^{i-}(k+1) \\&- \boldsymbol{K}_i(k+1)\boldsymbol{S}_{ab}(k+1)\boldsymbol{K}_i(k+1)^\top
\end{split}
\label{equ::EKF_update_P_self}
\end{equation}
\begin{equation}
\begin{split}
\boldsymbol{P}^{+}_{ij}(k+1) &= \boldsymbol{P}^{-}_{ij}(k+1) \\&- \boldsymbol{K}_i(k+1)\boldsymbol{S}_{ab}(k+1)\boldsymbol{K}_j(k+1)^\top
\end{split}
\label{equ::EKF_update_P_cor}
\end{equation}
\begin{multline}
\boldsymbol{K}_i(k+1) = \\
\begin{cases}
\boldsymbol{0}_{2} & \text{if no measurement}, \\
\begin{aligned}
&(\boldsymbol{P}^{-}_{ia}(k+1)\boldsymbol{H}_{ab,a}^\top(k+1) \\&+ \boldsymbol{P}^{-}_{ib}(k+1)\boldsymbol{H}_{ab,b}^\top(k+1))\boldsymbol{S}_{ab}^{-1}
\end{aligned} &\text{if}~a\xrightarrow{k+1} b,
\end{cases}
\label{equ::EKF_update_K_gain}
\end{multline}
\end{subequations}
To process multiple concurrent measurements, we use sequential updating (see~\mbox{\cite[page 103]{YB-PKW-XT:11}}). Let $\mathcal{V}_{\text{A}}(k)$ be the set of the robot that have taken a relative measurement with respect to other robots at timestep $k$. Let $\mathcal{V}_\text{\text{B}}^i(k)$ be the set of the landmark robots of robot $i\in\mathcal{V}_{\text{A}}(k)$.
Then, the total number of relative measurements at timestep $k$ is   $n_s(k)=\sum\nolimits_{i=1}^{|\mathcal{V}_{\text{A}}(k)|}|\mathcal{V}_\text{\text{B}}^i(k)|$. In sequential updating, the measurements are processed one by one, starting with using the first measurement to update the predicted estimate and error covaraince matrix, and proceeding with next measurement to update the current updated state estimate and error measurements. That is, we let
     $\hat{\boldsymbol{x}}^{i+}(k+1,0)=\hat{\boldsymbol{x}}^{i-}(k+1)$,
  $\boldsymbol{P}^{i+}(k+1,0)=\boldsymbol{P}^{i-}(k+1)$, $i\in\mathcal{V}$, and
  $\boldsymbol{P}_{i,l}^{+}(k+1,0)=\boldsymbol{P}_{il}^{i-}(k+1)$ for
  $l\in\mathcal{V}\backslash\{i\}$. Then, the sequential updating proceeds  (starting at $j=1$),
\begin{equation}
    \begin{aligned}
    &\text{for}~ a\in\mathcal{V}_{\text{A}}(k+1),
   \\
   &\quad\text{for}~ b\in\mathcal{V}^a_{\text{B}}(k+1),\quad\quad\quad\\
   &\quad\quad\hat{\boldsymbol{x}}^{i+}(k+1,j) \,\leftarrow \text{r.h.s of }\eqref{equ::EKF_update_x}, \\
   &\quad\quad\boldsymbol{P}^{i+}(k+1,j)\leftarrow\text{ r.h.s of  } \eqref{equ::EKF_update_P_self},\\
   &\quad\quad\boldsymbol{P}_{il}^{+}(k+1,j)\,\leftarrow\text{ r.h.s of }\eqref{equ::EKF_update_P_cor},\\
      &\quad~~~ j\leftarrow j+1,
    \end{aligned}
\label{eq::sequ-updt-CL-step}
\end{equation}
where at each $j$, $\hat{\boldsymbol{x}}^{i-}(k+1)$, $\boldsymbol{P}^{i-}(k+1)$, and  $\boldsymbol{P}_{il}^{-}(k+1)$ in~\eqref{eq::Joint_update} are replaced by, respectively, $\hat{\boldsymbol{x}}^{i+}(k+1,j-1)$, $\boldsymbol{P}^{i+}(k+1,j-1)$, and  $\boldsymbol{P}_{il}^{+}(k+1,j-1)$. Then the final update at timestep {$k+1$} is
  $\hat{\boldsymbol{x}}^{i+}(k+1)=\hat{\boldsymbol{x}}^{i+}(k+1,n_s(k+1))$,
  $\boldsymbol{P}^{i+}(k+1)=\boldsymbol{P}^{i+}(k+1,n_s(k+1))$, and
  $\boldsymbol{P}_{il}^{+}(k+1)=\boldsymbol{P}_{il}^{+}(k+1,n_s(k+1))$,
  $i\in\mathcal{V}$, $l\in\mathcal{V}\backslash\{i\}$.  

In what follows, we let $\boldsymbol{P}_c^{+}$ be the updated joint covariance of the team after all the concurrent relative measurements are processed, i.e., 
\begin{equation}\label{eq::join_P_c}
	\boldsymbol{P}_c^{+}(k)= \begin{bmatrix}
	\boldsymbol{P}^{1+}(k) & \cdots & \boldsymbol{P}_{1N}^{+}(k) \\
	\vdots &\ddots&\vdots\\
	\boldsymbol{P}_{N1}^{+}(k) & \cdots & \boldsymbol{P}^{N+}(k)
	\end{bmatrix},
\end{equation}
where $\boldsymbol{P}^{i+}(k)=\mathsf{E}[\tilde{\boldsymbol{x}}^{i+}(k)\tilde{\boldsymbol{x}}^{i+}(k)^\top]$ and  $\boldsymbol{P}_{ij}^{+}(k)=\mathsf{E}[\tilde{\boldsymbol{x}}^{i+}(k)\tilde{\boldsymbol{x}}^{j+}(k)^\top]$.

\subsection{Objective statement}
The joint EKF based CL algorithm described by~\eqref{equ::EKF_prop_x},~\eqref{equ::EKF_prop_P}, and~\eqref{eq::sequ-updt-CL-step} can be implemented in a decentralized manner following the methods proposed in \cite{roumeliotis2002distributed,SSK-SF-SM:16,kia2018serverassisted,luft2018recursive}. For example, an approach based on the interim-master decentralized CL IMDCL  
algorithm of~\cite{SSK-SF-SM:16} is as follows. Every robot maintains and propagates its own state estimate~\eqref{equ::EKF_prop_x} and corresponding error covariance~\eqref{equ::EKF_prop_Pi}. Every robot also stores a local copy of the cross-covariance components of the joint covariance matrix~\eqref{eq::join_P_c}\footnote{We note here that for robots with motion model of~\eqref{equ::EKF_sys_motion}
the locally stored variables of the decentralized CL algorithm of~\cite{SSK-SF-SM:16} at each robot $i\in\mathcal{V}$ are $\vect{\phi}^i(k)=\boldsymbol{I}_2$, $\vect{\Pi}^i_{ij}(k)=\boldsymbol{P}_{ij}^{-}(k+1)$ and $\vect{\Pi}^i_{ij}(k+1)=\boldsymbol{P}_{ij}^{+}(k+1)$, $j\in\mathcal{V}\backslash\{i\}$, for any $k\in\mathbb{Z}_{\geq0}.$}. Then, if at any time $k$ a robot $a$ takes a relative measurement with respect to another team member $b$, it can acquire the local a priori estimates $(\hat{\boldsymbol{x}}^{b-}(k),\boldsymbol{P}^{b-}(k))$ from $b$ and compute $\boldsymbol{K}_i$, $i\in\mathcal{V}$ and $\boldsymbol{S}_{ab}$ locally and~broadcast it to the rest of the team so every robot can update their estimates according to~\eqref{eq::Joint_update}. 
The intrinsic information exchange process of this decentralized operation leads to a stringent requirement on network connectivity and channel capacity. The communication operation also leads to energy usage at all networked robots, further consuming the limited on-board resources at the side of each robot. In such a context, it is naturally desirable to reduce the communication costs for a slightly reduced but still acceptable localization accuracy to achieve a globally better performance-resource trade-off. 

Following a sequential processing procedure, we can see from~\eqref{equ::EKF_update_P_self} that the more relative measurements are processed at each time $k$, the more reduction in the error covariance of the robots is achieved. However, to process each relative measurement there is a need for a robot-to-robot communication in the team. To create a balanced trade-off between localization accuracy and  communication utilization, one can restrict every robot $i\in\mathcal{V}$ to choose only $q^i$ number of landmark robots to take relative measurements from  out of all the possibilities. To achieve best localization with this constraint, the robots should choose their landmarks wisely  such that the total uncertainty in $\boldsymbol{P}_c^+(k+1)$ is minimized. This choice can be made by solving the following optimization problem,  
	\begin{align}\label{eq::joint_opt}
&	(\mathcal{V}_{\text{B}}^1(k+1),\dots,\mathcal{V}_{\text{B}}^N(k+1))=\underset{}{\rm argmin} \, {\rm det}(\boldsymbol{P}^+_c(k+1))\nonumber
	\\
	&\qquad\qquad\qquad\qquad\text{s.t.}\quad |\mathcal{V}_{\text{B}}^i(k+1)|\leq q^i, ~~i\in\mathcal{V}.
	\end{align}
 The choice of the determinant of the collective covariance, ${\rm det}(\boldsymbol{P}_c^+(k+1))$, as the objective function is motivated by the fact that the determinant of covariance is directly linked to the differential entropy, which describes the volume of uncertainty ellipse \cite{tzoumas2017scheduling}. For each robot, there are at most $N\!-\!1$ inter-robot relative measurements available, i.e., $|\mathcal{V}_{\text{B}}^i(k)|\!\leq\! N-1$. To choose at most $q^i\in\mathbb{Z}_{\geq 1}$ measurements for each robot  $i\in\mathcal{V}$ from all the possible ones is a classical NP-hard sensor-selection problem \cite{tzoumas2017scheduling,zhang2017sensor}.
Also, given that $\boldsymbol{P}^+_c(k+1)$ is the joint covariance matrix of all the robots, obtaining a decentralized solution for the optimization problem~\eqref{eq::joint_opt} is challenging. To arrive at a tractable decentralized solution, one can replace~\eqref{eq::joint_opt} with the following suboptimal landmark selection for each robot $i\in\mathcal{V}$:
	\begin{align}\label{eq::prob_def}
	\mathcal{V}_{\text{B}}^i(k+1)&=\underset{}{\rm argmin} \, \det(\boldsymbol{P}^+_{c,i}(k+1))~\text{s.t.}\nonumber
	\\
	&~~|\mathcal{V}_{\text{B}}^i(k+1)|\leq q^i.
	\end{align}
where $\boldsymbol{P}^+_{c,i}(k+1)$ is the updated joint covariance matrix of the network if we only use the measurements taken by robot $i$ from landmark robots  $\mathcal{V}_{\text{B}}^i(k+1)$. Problem~\eqref{eq::prob_def} is still an NP-hard problem. Also, since $\boldsymbol{P}^+_{c,i}(k+1)$ depends on the local covariance matrices of the other robots in the team, robot $i\in\mathcal{V}$ needs to communicate and collaborate with the rest of the team to solve~\eqref{eq::prob_def}. When the cost function in~\eqref{eq::prob_def} is replaced by equivalent form of $\text{log}\det(\boldsymbol{P}^+_{c,i}(k+1))$
the resulted equivalent problem becomes a sub-modular optimization problem, for which suboptimal greedy solutions with polynomial time computational complexity are explored in~\cite{tzoumas2017scheduling} and \cite{tzoumas2016nearoptimal}. However, these solutions suffer from high communication cost because each robot needs to have access to the joint covariance matrix $\boldsymbol{P}_c^-(k+1)$ whose diagonal elements $\{\boldsymbol{P}^{i-}(k+1)\}_{i\in\mathcal{V}}$ are maintained by respective robot $i\in\mathcal{V}$.

\textbf{Objective 1 (Measurement scheduling under restricted communication)}: the objective in this paper is to obtain a suboptimal solution with a polynomial time computational complexity for~\eqref{eq::prob_def}, where each robot chooses its own $q^i$ set of the landmark robots to take relative measurements from locally. In developing our solution, we impose the condition that t
The only information available to each robot $i\in\mathcal{V}$ to obtain its own suboptimal solution is robot $i$'s own state estimate and its corresponding error covariance along with a copy of robot $i$'s state estimate cross-covariances with respect to other robots in the team, i.e., $\boldsymbol{x}^{i+}(k)$, $\boldsymbol{P}^{i+}(k)$, and  $\{\boldsymbol{P}^{+}_{ij}(k)\}_{ j\in\mathcal{V}\backslash\{i\}}$.
That is, each robot should obtain its suboptimal landmark selection solution by using its own local data and without any cooperation with the other robots in the~team. \boxend

\section{A suboptimal decentralized measurement scheduling}
\label{sec::mea_sche}
In this section, we provide a novel solution that meets Objective 1. Our solution relies on obtaining an upper-bound on $\det(\boldsymbol{P}_c^{+})$, where $\boldsymbol{P}_c^{+}$ is the updated joint covariance matrix due to relative measurement $a\rightarrow b$, that depends only on the locally available variables at robot $a$. 

\subsection{Upper Bound on Determinant of joint Covariance}
In what follows, we derive an upper bound of the state-dependent objective variable ${\rm det}(\boldsymbol{P}_c^+(k+1))$ so that the optimization problem becomes tractable. First, we note that in the propagation stage, the incremental covariance $\boldsymbol{Q}^i(k)$ in \eqref{equ::EKF_prop_P_Q} can be upper bounded by constant matrix $\check{\boldsymbol{Q}}^i$ as follows
\begin{equation*}
	\boldsymbol{Q}^i(k) \leq \check{\boldsymbol{Q}}^i = (\delta t)^2 {\rm max}(\sigma_{v^i}^2,(v_{max}^i)^2\sigma_{\phi^i}^2) \boldsymbol{I}_2,
\end{equation*}
where $v_{max}^i$ is the maximal linear velocity of robot $i$. Here, we use the fact that the rotation matrix $\boldsymbol{C}({\phi}_m^i(k))$ satisfies  $\boldsymbol{C} ({\phi}_m^i(k))\leq \boldsymbol{I}_2$. Then, given~\eqref{equ::EKF_prop_P} the collective covariance matrix should satisfy 
\begin{equation*}
	\boldsymbol{P}_c^-(k+1) \leq \boldsymbol{P}_c^+(k) + \check{\boldsymbol{Q}}_c
\end{equation*}
with $\check{\boldsymbol{Q}}_c = {\rm diag}(\check{\boldsymbol{Q}}^1,\cdots,\check{\boldsymbol{Q}}^N)$.
By setting $\check{\boldsymbol{P}}_c^-(k+1) = \check{\boldsymbol{P}}_c^+(k) + \check{\boldsymbol{Q}}_c$ and $\check{\boldsymbol{P}}_c^+(0) = \boldsymbol{P}_c^+(0)$
, we have $\boldsymbol{P}_c^-(k+1)\leq \check{\boldsymbol{P}}_c^-(k+1)$,
which according to~\cite[Corollary 18.1.8]{harville1997matrix}  guarantees that 
${\rm det}(\boldsymbol{P}_c^-(k+1)) \leq {\rm det}(\check{\boldsymbol{P}}_c^-(k+1))$.

Next, we derive an upper bound on the determinant of updated joint covariance corresponding to the relative measurement $a\overset{k+1}{\longrightarrow}b$.
Using standard manipulations the joint updated covariance matrix described by \eqref{equ::EKF_update_P_cor} and  \eqref{equ::EKF_update_K_gain}, in its information form reads as
\begin{equation}
\begin{split}
	&(\boldsymbol{P}_c^+(k+1))^{-1} = (\boldsymbol{P}_c^-(k+1))^{-1} \\&\quad+ \boldsymbol{H}_{c}^\top(k+1) (\boldsymbol{R}_{ab}(k+1))^{-1} \boldsymbol{H}_{c}(k+1),
\end{split}
\label{equ::EKF_update_P_c_IF_0}
\end{equation}
where $\boldsymbol{H}_{c} = \begin{bmatrix}
\overbrace{\boldsymbol{0}_2}^1 & \cdots & \overbrace{\boldsymbol{H}_{ab,a}}^{a} & \cdots & \overbrace{\boldsymbol{H}_{ab,b}}^{b} & \cdots & \overbrace{\boldsymbol{0}_2}^N
\end{bmatrix}$ and $\boldsymbol{R}_{ab} = \boldsymbol{R}_{z_{ab}} + \boldsymbol{R}_{\phi_m^{ab}}$.
Substituting \eqref{equ::EKF_update_S_R} into \eqref{equ::EKF_update_P_c_IF_0} results in \cite{mourikis2006performance}
\begin{equation}
\begin{split}
&(\boldsymbol{P}_c^+(k+1))^{-1} = (\boldsymbol{P}_c^-(k+1))^{-1} \\&+ \check{\boldsymbol{H}}_{c}^\top(k+1) (\boldsymbol{R}_{c,ab}(k+1))^{-1} \check{\boldsymbol{H}}_{c}(k+1),
\end{split}
\label{equ::EKF_update_P_c_IF_1}
\end{equation}
where $\check{\boldsymbol{H}}_{c} = \begin{bmatrix}
\overbrace{\boldsymbol{0}_2}^1 & \cdots & \overbrace{-\boldsymbol{I}_2}^{a} & \cdots & \overbrace{\boldsymbol{I}_2}^{b} & \cdots & \overbrace{\boldsymbol{0}_2}^N
\end{bmatrix}$ and $\boldsymbol{R}_{c,ab} = \sigma_{\rho^a}^2 \boldsymbol{I}_2 - \boldsymbol{D}_{ab} {\rm diag}(\frac{\sigma_{\rho^a}^2}{\hat{\rho}_{ab}^2})\boldsymbol{D}_{ab}^\top +\sigma_{\theta^a}^2\boldsymbol{D}_{ab} \boldsymbol{D}_{ab}^\top + \sigma_{\phi^a}^2\boldsymbol{D}_{ab}\boldsymbol{1}_2\boldsymbol{D}_{ab}^\top.$
Here, 
$\boldsymbol{D}_{ab} = {\rm diag}(\boldsymbol{J}
( \hat{\boldsymbol{x}}^{b-}(k) - \hat{\boldsymbol{x}}^{a-}(k)))$ is the diagonal matrix with diagonal elements from vector $\boldsymbol{J}
( \hat{\boldsymbol{x}}^{b-}(k) - \hat{\boldsymbol{x}}^{a-}(k))$ and $\hat{\rho}^2_{ab} = (\hat{\boldsymbol{x}}^{b-}(k) - \hat{\boldsymbol{x}}^{a-}(k))^\top (\hat{\boldsymbol{x}}^{b-}(k) - \hat{\boldsymbol{x}}^{a-}(k))$.
The orientation-dependent terms are now cancelled out with $\check{\boldsymbol{H}}_{c}$ being constant.
Considering the physical constraints on the measurement ranges of the robots, $\boldsymbol{R}_{c,ab}$ can be upper bounded by constant matrix $\check{\boldsymbol{R}}_{c,a}$ as $\boldsymbol{R}_{c,ab}(k) \leq \check{\boldsymbol{R}}_{c,a} =  r_{c,a}\boldsymbol{I}_2$
where $r_{c,a} = (\sigma_{\rho^a}^2 + \sigma_{\phi^a}^2\rho_{max,a}^2 + \sigma_{\theta^a}^2\rho_{max,a}^2)$ and
$\rho_{max,a}$ is the maximal sensing range for robot $a$ (see~ \cite[Appendix I]{mourikis2006performance}).
Substituting the inequality into \eqref{equ::EKF_update_P_c_IF_1}, we have
\begin{equation}\label{eq::PC+_upperB}
\boldsymbol{P}_c^+(k+1) \leq \check{\boldsymbol{P}}_c^+(k+1),
\end{equation}
where
\begin{equation}\label{eq::P_c_upper_bound}
    \check{\boldsymbol{P}}_c^+(k+1)=\left( (\boldsymbol{P}_c^-(k+1))^{-1}+ \frac{1}{r_{c,a}}\check{\boldsymbol{H}}_{c}^\top \check{\boldsymbol{H}}_{c} \right)^{-1}.
\end{equation}

Then, according to~\cite[Corollary 18.1.8]{harville1997matrix} it is guaranteed that
${\rm det}(\boldsymbol{P}_c^+(k+1)) \leq {\rm det}(\check{\boldsymbol{P}}_c^+(k+1))$.
To sum up, we construct the upper bound of covariance $\check{\boldsymbol{P}}_c^+$
such that ${\rm det}({\boldsymbol{P}}_c^+(k)) \leq {\rm det}(\check{\boldsymbol{P}}_c^+(k))$ always holds given that $\check{\boldsymbol{P}}_c^+(0) = \boldsymbol{P}_c^+(0)$. Using the upper bounds established so far, we make the following statement. In what follows, to simplify the notation, we drop the time argument.

\medskip
\begin{theorem}\label{thm::main}
Consider the joint updated covariance matrix $\boldsymbol{P}_c^+$ due to the relative measurement $a\rightarrow b$. Let $\check{\boldsymbol{P}}^{-}_{ab}$
be the block matrix in the $a^{th}$ row and $b^{th}$ column of $\check{\boldsymbol{P}}_c^-$. Then,
\begin{equation}
\begin{split}
& {\rm det}({\boldsymbol{P}}_c^+) \leq 
\\& \frac{{\rm det}(\check{\boldsymbol{P}}_c^-)}{
	1 + r_{c,a}^{-1}{\rm tr}( \check{\boldsymbol{P}}_{aa}^- \!+\! \check{\boldsymbol{P}}_{ba}^- (\check{\boldsymbol{P}}_{aa}^-)^{-1}
	\check{\boldsymbol{P}}_{ab}^- \!-\! \check{\boldsymbol{P}}_{ab}^- \!-\! \check{\boldsymbol{P}}_{ba}^-) },
\end{split}
\label{equ::main_theorem}
\end{equation} 
and
\begin{equation}\label{equ::main_theorem2}
    {\rm tr}( \check{\boldsymbol{P}}_{aa}^- + \check{\boldsymbol{P}}_{ba}^- (\check{\boldsymbol{P}}_{aa}^-)^{-1} \check{\boldsymbol{P}}_{ab}^- - \check{\boldsymbol{P}}_{ab}^- - \check{\boldsymbol{P}}_{ba}^-) 
\geq 0.
\end{equation}
\end{theorem}
\medskip
\begin{IEEEproof}
Given that
$\check{\boldsymbol{P}}_c^+ = ( (\check{\boldsymbol{P}}_c^-)^{-1}+
r_{c,a}^{-1} \check{\boldsymbol{H}}_{c}^\top \check{\boldsymbol{H}}_{c} )^{-1}$, using some algebraic manipulation and invoking~\cite[Corollary 18.1.2]{harville1997matrix}, we can write
${\rm det}(\check{\boldsymbol{P}}_c^+) = 
	\frac{1}{{\rm det}\left( (\check{\boldsymbol{P}}_c^-)^{-1}+ r_{c,a}^{-1} \check{\boldsymbol{H}}_{c}^\top \check{\boldsymbol{H}}_{c} \right)}= $
	$\frac{{\rm det}(\check{\boldsymbol{P}}_c^-)}{
		{\rm det}\left( \boldsymbol{I}_{2N}+
		 r_{c,a}^{-1}
		\check{\boldsymbol{P}}_c^- \check{\boldsymbol{H}}_{c}^\top \check{\boldsymbol{H}}_{c}\right)}= $
		$\frac{{\rm det}(\check{\boldsymbol{P}}_c^-)}{
	{\rm det}\left( \boldsymbol{I}_{2}+
		 r_{c,a}^{-1} \check{\boldsymbol{H}}_{c} \check{\boldsymbol{P}}_c^- \check{\boldsymbol{H}}_{c}^\top \right)}$.
		 Then, by virtue of Lemma \ref{lem::det_tr_ineq_0} we~obtain
\begin{equation}\label{equ::det_P_c_ineq}
	{\rm det}(\check{\boldsymbol{P}}_c^+) \leq \frac{{\rm det}(\check{\boldsymbol{P}}_c^-)}{		1 + {\rm tr}( r_{c,a}^{-1} \check{\boldsymbol{H}}_{c} \check{\boldsymbol{P}}_c^- \check{\boldsymbol{H}}_{c}^\top)}.
\end{equation}

Since $\begin{bmatrix}
\check{\boldsymbol{P}}_{aa}^- & \check{\boldsymbol{P}}_{ab}^-\\
\check{\boldsymbol{P}}_{ba}^- & \check{\boldsymbol{P}}_{bb}^-
\end{bmatrix}\geq 0$,
using Lemma \ref{lem::det_tr_ineq_1} one can write ${\rm tr}( \check{\boldsymbol{P}}_{aa}^- + \check{\boldsymbol{P}}_{bb}^- - \check{\boldsymbol{P}}_{ab}^- - \check{\boldsymbol{P}}_{ba}^-) 
\geq {\rm tr}( \check{\boldsymbol{P}}_{aa}^- + \check{\boldsymbol{P}}_{ba}^- (\check{\boldsymbol{P}}_{aa}^-)^{-1} \check{\boldsymbol{P}}_{ab}^- - \check{\boldsymbol{P}}_{ab}^- - \check{\boldsymbol{P}}_{ba}^-) 
\geq 0$.
Then, using 
 $   \check{\boldsymbol{H}}_{c} \check{\boldsymbol{P}}_c^- \check{\boldsymbol{H}}_{c}^\top=
\check{\boldsymbol{P}}_{aa}^- - \check{\boldsymbol{P}}_{ab}^- -
\check{\boldsymbol{P}}_{ba}^- + \check{\boldsymbol{P}}_{bb}^-.$  to  expand the denominator of the right hand side expression in~\eqref{equ::det_P_c_ineq}, we have ${\rm tr}( r_{c,a}^{-1} \check{\boldsymbol{H}}_{c} \check{\boldsymbol{P}}_c^- \check{\boldsymbol{H}}_{c}^\top)
=r_{c,a}^{-1}{\rm tr}( \check{\boldsymbol{P}}_{aa}^- + \check{\boldsymbol{P}}_{bb}^- - \check{\boldsymbol{P}}_{ab}^- - \check{\boldsymbol{P}}_{ba}^-)
\geq r_{c,a}^{-1}{\rm tr}( \check{\boldsymbol{P}}_{aa}^- + \check{\boldsymbol{P}}_{ba}^- (\check{\boldsymbol{P}}_{aa}^-)^{-1}
\check{\boldsymbol{P}}_{ab}^- - \check{\boldsymbol{P}}_{ab}^- - \check{\boldsymbol{P}}_{ba}^-) \geq 0.$
Therefore, it follows from~\eqref{equ::det_P_c_ineq} that
${\rm det}(\check{\boldsymbol{P}}_c^+) \leq \frac{{\rm det}(\check{\boldsymbol{P}}_c^-)}{
	1 + r_{c,a}^{-1}{\rm tr}( \check{\boldsymbol{P}}_{aa}^- + \check{\boldsymbol{P}}_{ba}^- (\check{\boldsymbol{P}}_{aa}^-)^{-1}
	\check{\boldsymbol{P}}_{ab}^- - \check{\boldsymbol{P}}_{ab}^- - \check{\boldsymbol{P}}_{ba}^-) }.
$ The proof is then completed due to~\eqref{eq::PC+_upperB}.
\end{IEEEproof}


\medskip
\subsection{Scheduling Using Locally Stored  Covariance and Cross-covariance Matrices}
Our measurement scheduling algorithm design to meet Objective 1 relies on the results given by Theorem~\ref{thm::main}. We note that in a sequential scheduling update procedure, each relative measurement $a\overset{k}{\longrightarrow}b$ has separate effect on the determinant of the joint covariance matrix. We can then make scheduling individually for each robot for uncertainty minimization. We note that since the denominator of right hand side in \eqref{equ::main_theorem} is always positive, its maximization leads to the reduction of the upper bound on ${\rm det}({\boldsymbol{P}}_c^+)$.
Given this observation, we  propose Algorithm~\ref{alg::subopt} below as a greedy landmark selection procedure at each robot $i$. We let $\mathcal{D}^i(k)$ be the set of robots that robot $i\in\mathcal{V}$ detects in its measurement zone at time $k$ (e.g., by detecting the AR tags in the image taken by its camera~\cite{niekum2013ar_track_alvar}). Robot $i$ uses Algorithm~\ref{alg::subopt} to decide which $q^i$ number of robots it should take relative measurement from (e.g., by processing its image further to extract the relative pose of only those $q^i$ selected landmark robots~\cite{niekum2013ar_track_alvar}).
\begin{algorithm}
	\textbf{Input:} $\mathcal{D}^i(k)$, covariance $\boldsymbol{P}^{i-}(k)$,  cross-covariances 
	$\{\boldsymbol{P}_{ij}^-(k)\}_{j\in\mathcal{D}^i(k)}$, constraint on the number of measurements $q^i$, constant $r_{c,i}$ \\
	\textbf{Output: } the identification of the $\min\{q^i,|\mathcal{D}^i(k)|\}$ landmark robots for robot $i\in\mathcal{V}$ 
	
	\begin{algorithmic}
		\IF{ $q^i< |\mathcal{D}^i(k)|$}
		{
		\FOR{$j\in\mathcal{D}^i(k) $}
		\STATE { ~~$\boldsymbol{J}_{ij} = r_{c,i}^{-1}{\rm tr}( \check{\boldsymbol{P}}_{ii}^- + \check{\boldsymbol{P}}_{ji}^- (\check{\boldsymbol{P}}_{ii}^-)^{-1}
		\check{\boldsymbol{P}}_{ij}^- - 
		\check{\boldsymbol{P}}_{ij}^- -\check{\boldsymbol{P}}_{ji}^-)$}  
		\ENDFOR
		\STATE{find the largest $q^i$ elements of the $\{\boldsymbol{J}_{ij}\}_{j\in\mathcal{D}^i(k)}$ scalar set and output the corresponding subscripts $j$}
		}
		\ELSE{\RETURN{$\mathcal{D}^i(k)$}}
		\ENDIF
	\end{algorithmic}
	\caption{Landmark selection at robot $i$}
	\label{alg::subopt}
\end{algorithm}

To implement Algorithm~\ref{alg::subopt}, each robot $i\in\mathcal{V}$ uses its local information to compute  $\boldsymbol{J}_{ij} = r_{c,i}^{-1}{\rm tr}( \check{\boldsymbol{P}}_{ii}^- + \check{\boldsymbol{P}}_{ji}^- (\check{\boldsymbol{P}}_{ii}^-)^{-1}
\check{\boldsymbol{P}}_{ij}^- - \check{\boldsymbol{P}}_{ij}^- - \check{\boldsymbol{P}}_{ji}^-)$, ~$j\in\mathcal{D}^i(k)$. Then, it chooses $q^i$ largest values of $\boldsymbol{J}_{ij}$,~$j\in\mathcal{D}^i(k)$, as its landmark robots. 
Of course, if $|\mathcal{D}^i(k)|<q^i$ then robot $i$ uses $\mathcal{D}^i(k)$ as its landmark~robots. 
We note here that 
since the robot-wise incremental covariance in propagation stage $\boldsymbol{Q}^i(k)$ is dependent on the velocity input and is thus known to each robot locally, we can discard $\check{\boldsymbol{Q}}^i(k)$ in scheduling.
That is, when we implement Algorithm~\ref{alg::subopt}, we can use  $\check{\boldsymbol{P}}_{ii}^-(k) = \boldsymbol{P}^{i-}(k)$ and $\check{\boldsymbol{P}}_{ij}^-(k) = \boldsymbol{P}_{ij}^-(k)$, $i\in\mathcal{V}$ and~$j\in\mathcal{D}^i(k)$. 

Algorithm~\ref{alg::subopt} is a greedy landmark selection heuristic that works based on minimizing an upper-bound on the total uncertainty of the team. Even though this algorithm does not have rigorous performance guarantees, our numerical examples in the proceeding section shows that the performance of CL algorithm implementing Algorithm~\ref{alg::subopt} is comparable to the CL algorithm that uses the landmark selection algorithm of~\cite{tzoumas2017scheduling}, which comes with a known optimality gap. This observation along with the properties highlighted in the remarks below makes Algorithm~\ref{alg::subopt} an appealing choice for operations with resource constrained robots.

\begin{rem}[Computational and communication cost of Algorithm~\ref{alg::subopt}]{\rm 
First, we observe that implementing Algorithm~\ref{alg::subopt} poses no communication overhead on the robots, i.e., to carry out Algorithm~\ref{alg::subopt} robots do not need to communicate with each other. 
In contrast, in the landmark selection algorithm of~\cite{tzoumas2017scheduling} each robot $i\in\mathcal{V}$ needs to know the local covariance matrix of all the other teammates whether they are in $\mathcal{D}^i$ or not, i.e.,~\cite{tzoumas2017scheduling} requires all-to-all communication for landmark selection at each timestep.
Algorithm~\ref{alg::subopt} is a  numerically efficient procedure, as well. It only computes and ranks $\{\boldsymbol{J}_{ij}\}_{j\in\mathcal{D}^i}$, which are at most $N-1$ scalars.
The complexity of computing all $\boldsymbol{J}_{ij}$ mainly comes from the matrix multiplication and inversion, which is $O(2^{2.4}(N-1)) \simeq O(N)$ in total \cite{press2007numerical}.
Ranking and selecting the largest $q^i$ ones requires time complexity of $O(N\text{log}N)$.
Therefore the total complexity of our proposed suboptimal scheduling algorithm is $O(N\text{log}N)$ per robot, while the suboptimal method in \cite{tzoumas2017scheduling} always has complexity of $O((N-1)q^i(2N)^{2.4})\simeq O(q^iN^{3.4})$ for each robot,
regardless of the size of $\mathcal{D}^i$, i.e., the number of the robots in the measurement zone. The communication and computational costs after the landmark selection for a CL algorithm implementing Algorithm~\ref{alg::subopt} and the one implementing the landmark selection algorithm of~\cite{tzoumas2017scheduling} are the same since the scheduling does not change the measurement model and the update steps.
}\boxend
\end{rem}

\begin{rem}[No further restriction on the team beyond those imposed by the adopted decentralized CL]{\rm
 We note here that to establish the upper bound~\eqref{equ::main_theorem}, we made no assumptions about the type of the robots or the number or quality of the relative measurements. Moreover, note that Algorithm~\ref{alg::subopt} allows each robot to choose its constraint on the number of the relative measurements it wants to pick, $q^i$, and change its choice based on the status of its available resources at each~time. Therefore, Algorithm~\ref{alg::subopt} can be applied to teams of heterogeneous robots moving on a flat surface. 
Additionally, since the upper bound~\eqref{equ::main_theorem} has no direct dependency on the size of the team, Algorithm~\ref{alg::subopt} can be implemented in operations that the size of the team changes over time due to robots leaving (e.g., due to failure) or joining the operation. In fact, the change in the team size is of importance for the integrity of the CL algorithm rather than the landmark selection Algorithm~\ref{alg::subopt}. 
In case of the IMDCL algorithm, as stated in~\cite{SSK-SF-SM:16}, IMDCL is robust to permanent agent dropouts from the network. The operation only suffers from a processing cost until all agents become aware of the dropout. On the other hand, a new robot can join the team and participate in IMDCL as long as its addition is made known to all the teammates so that they initiate and maintain a cross-covaraince term corresponding to this new robot. 
}\boxend
 \end{rem}

\section{Numerical Simulation}
\label{sec::sim}
We demonstrate the efficacy of our scheduling algorithm by comparing its performance to that of the method of~\cite{tzoumas2017scheduling} 
and that of a random landmark selection
in two sets of simulation studies.
In our study, the noise variances of the robots, except for  $\sigma^i_{\phi}$, which we have selected, are taken from~\cite[Chapter 3]{leung2011utias,leung2012cooperative} as given in the table below.  The computer codes used to conduct our simulation studies are available at~\cite{code}.
\begin{center}
	\begin{tabular}{l c}
		\hline
		linear velocity measurement noise $\sigma_{\eta^i}$ & $2.253|v_{i,k}|$
	    \\ angular velocity measurement noise $\sigma_{\omega^i}$ & $0.587\text{ rad/s}$
		\\ distance sensing noise $ \sigma_{\rho^i}$ & $0.147 \text{ m}$ 
		\\ bearing sensing noise $\sigma_{\theta^i}$ & $0.1 \text{ rad}$
		\\ orientation measurement noise $\sigma_\phi^i$ & $0.0349 \text{ rad}$
		\\\hline
	\end{tabular}
	\end{center}

\begin{figure}
	\centering
	\includegraphics[width=2.9 in]{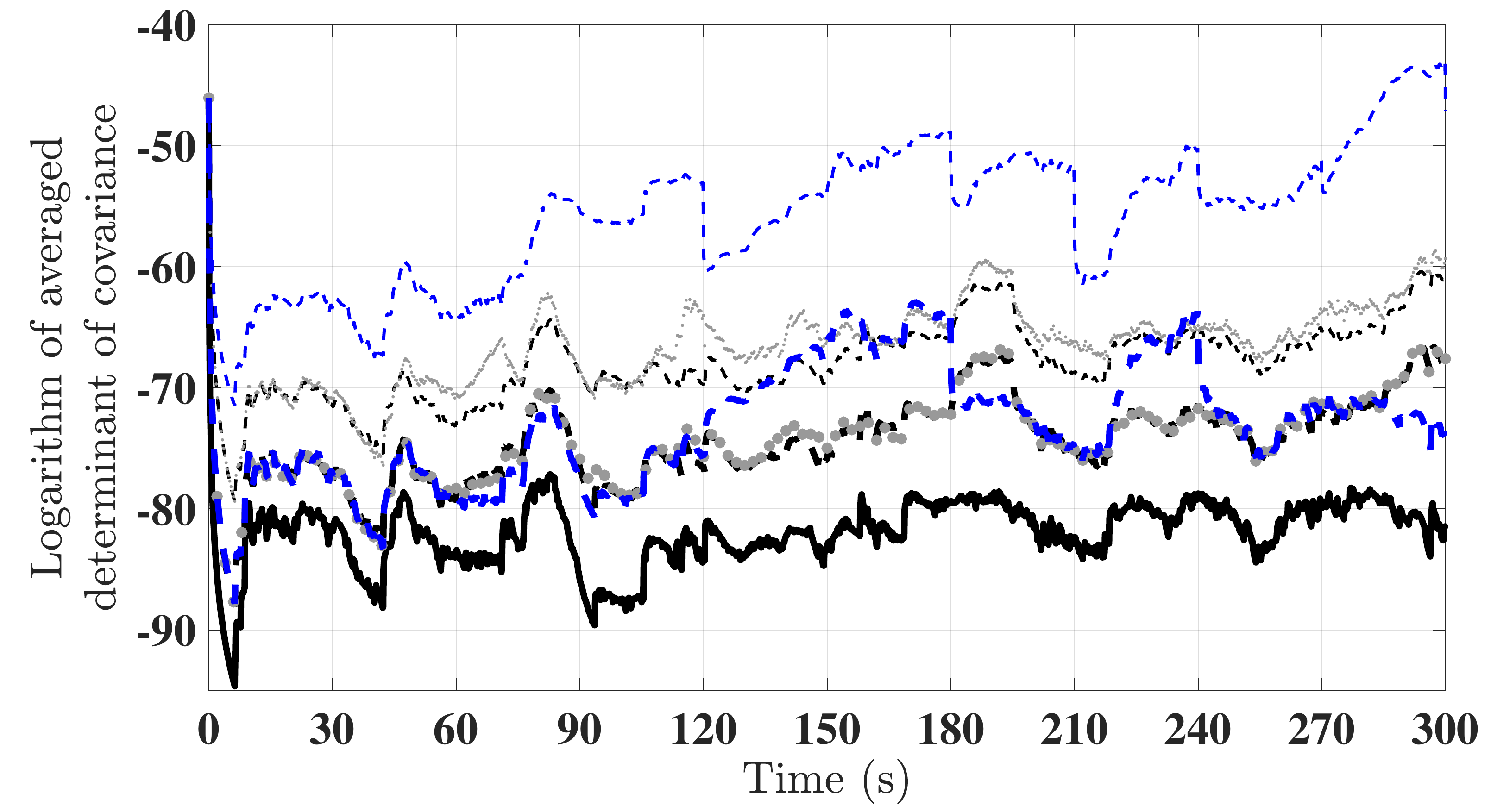}
	\caption{Logarithm of determinant of estimation covariance for the first simulation study: the solid thick line shows the result for $q^i = N-1=4$. 
	In the landmark selection scenarios when $q^i<N-1$,
	the blue dashed line shows the result for the random selection in which the landmark robots are randomly selected every $30$ seconds;
	the black dashed line shows the result when the suboptimal landmark selection algorithm of~\cite{tzoumas2017scheduling} is used;
	and finally the gray dotted line
	shows the result due to suboptimal landmark selection Algorithm~\ref{alg::subopt} proposed in this paper. 
	In all these cases, the thinner line corresponds to $q^i=1$ and the thicker one to $q^i=3$.}
	\label{fig::logdet_dataset}
\end{figure}

\emph{A simulation study based on a real-world dataset:}
We validate our proposed algorithm on the public UTIAS multi-robot cooperative
localization and mapping dataset~\cite{leung2011utias}, in which a team of $5$ robots move on a 2D flat surface in an indoor environment. The UTIAS dataset consists of 9 sub-datasets and each includes measurement data, odometry data and groundtruth position data of all the team members.
We use the first $300$ seconds from sub-dataset $7$. 
In this simulation, all 5 robots are allowed to take relative measurements with respect to any team members at each timestep.
Figure~\ref{fig::logdet_dataset} shows that if robots process all the relative measurements that are potentially available the best localization accuracy is achieved. However, as expected, when the robots are restricted to take $q^i$ number of relative measurements the localization performance drops. But, this is a trade-off for lower communication/computation cost. As seen in Fig.~\ref{fig::logdet_dataset}, our proposed measurement scheduling yields a comparable performance to that of the suboptimal scheduling solution of~\cite{tzoumas2017scheduling} (for both cases of $q^i=1$ and $q^i=3$).
However, we recall that our proposed algorithm does not require any inter-robot communication at scheduling stage as opposed to the suboptimal solution of~\cite{tzoumas2017scheduling}, which requires all-to-all communication at the time of measurement scheduling.
Figure~\ref{fig::logdet_dataset} shows that the random landmark selection delivers an inferior performance, especially in the case of $q^i = 1$, which well explains the necessity of solving the optimization problem~\eqref{eq::prob_def}.

\begin{table*}
	\centering
	\caption{Measurement time table for the Monte-Carlo study. 
	}
	\label{table::CL_time}
	\begin{tabular}{|c||c|c|c|c|c|c|c|c|c|}
		\hline
		Time (second) & [0 10] & (10 20] & (20 35] & (35 40] & (40 60] & (60 65] & (65 80] & (80 95] & (95 100] \\\hline
		\multirow{1}* {Robots allowed to take measurements}& none & 3, 5, 7, 9 & 2, 6, 8 & 1, 5, 7 & 3, 4, 6, 9 &5, 7&3, 6, 8 &1, 4, 9 &4, 6\\
        \hline
	\end{tabular}
\end{table*}
\emph{Monte-Carlo simulation}:
Next, we demonstrate the effectiveness of our proposed measurement scheduling algorithm through a Monte-Carlo simulation study for $9$ robots.
This simulation runs at $\delta t=0.1$ seconds.
In this simulation, robots move with constant linear velocity of $0.1\text { m/s}$ and rotational velocity of $0.1 \text{ rad/s}$. 
The initial conditions are $\boldsymbol{P}^i(0) = \text{diag}(0.01\text{m}^2,0.01\text{m}^2), \boldsymbol{P}_{ij}(0) = \boldsymbol{0}_2, i\in\mathcal{V}, j\in\mathcal{V}\backslash\{i\}$.
Initial estimated location $\hat{\boldsymbol{x}}^i(0)$ is generated according to the covariance, i.e. $\hat{\boldsymbol{x}}^i(0) \sim N(\boldsymbol{x}^i(0),\boldsymbol{P}^i(0)), i\in\mathcal{V}$.
Robots start from a 2-by-2 mesh lattice formation where distance between each vertices is $3$ meters. The true initial orientation $\phi^i(0)$ is uniformly drawn from $[0,2\pi)$.

\begin{figure}
	\centering
	\includegraphics[width=2.9 in]{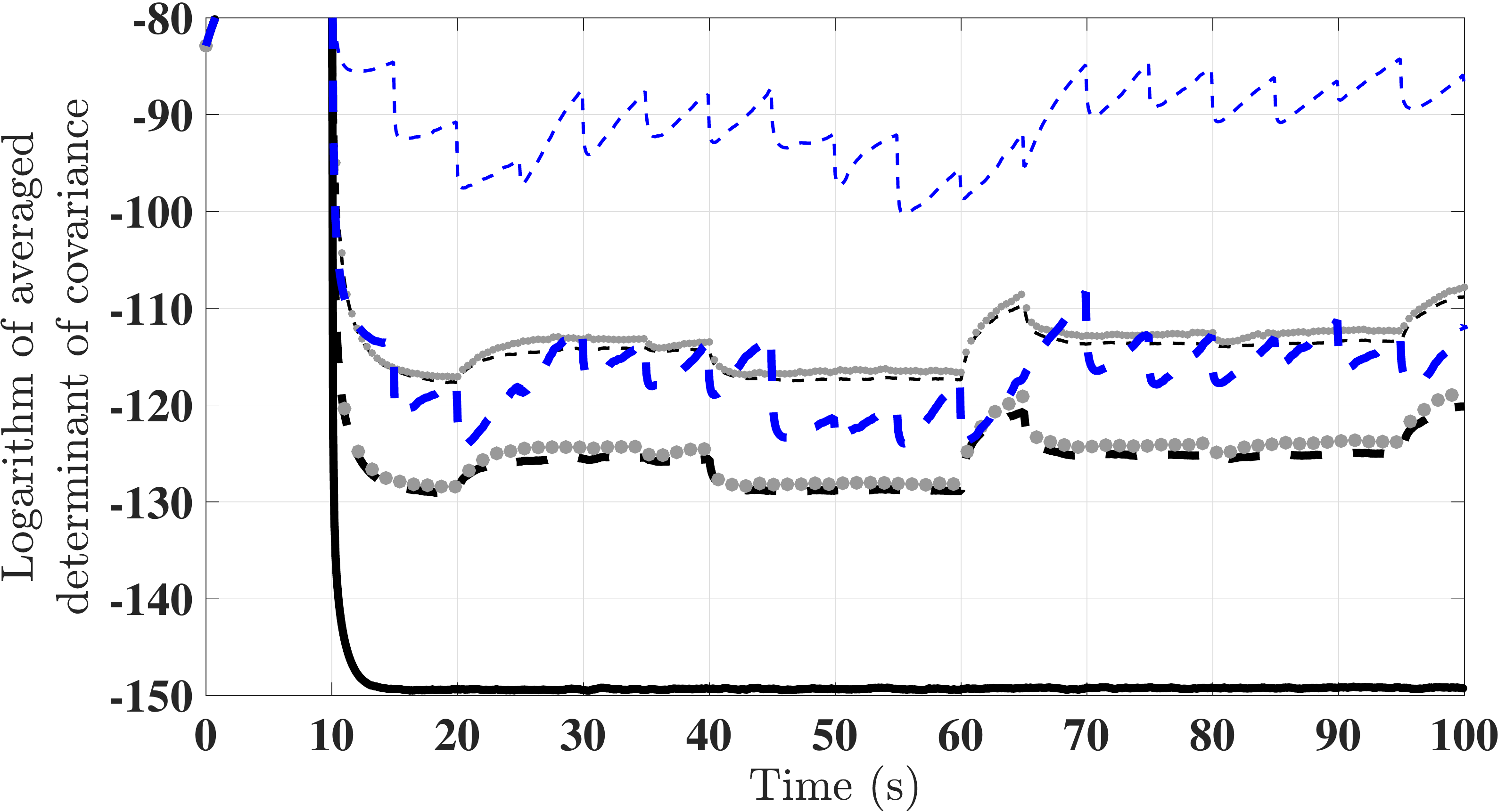}
    \vspace{-0.1 in}
	\caption{Logarithm of the averaged determinant of estimation covariance in the Monte Carlo study: the solid thick line shows the result when the robots that are allowed to take measurement (specified in Table~\ref{table::CL_time}) take relative measurement from all the $N-1=8$ landmark robots in their measurement zone. 
	In the landmark selection scenarios when $q^i<N-1$,
	the blue dashed line shows the result for the random selection in which the landmark robots are randomly selected every $5$ seconds;
	the black dashed line shows the result when the suboptimal landmark selection algorithm of~\cite{tzoumas2017scheduling} is used;
	and finally the gray dotted line
	shows the result due to suboptimal landmark selection Algorithm~\ref{alg::subopt} proposed in this paper. 
	In all these cases, the thinner line corresponds to $q^i=1$ and the thicker one to $q^i=3$.
	}
	\label{fig::logdet_MC}
\end{figure}

\begin{figure}
	\centering
	\includegraphics[width=2.9 in]{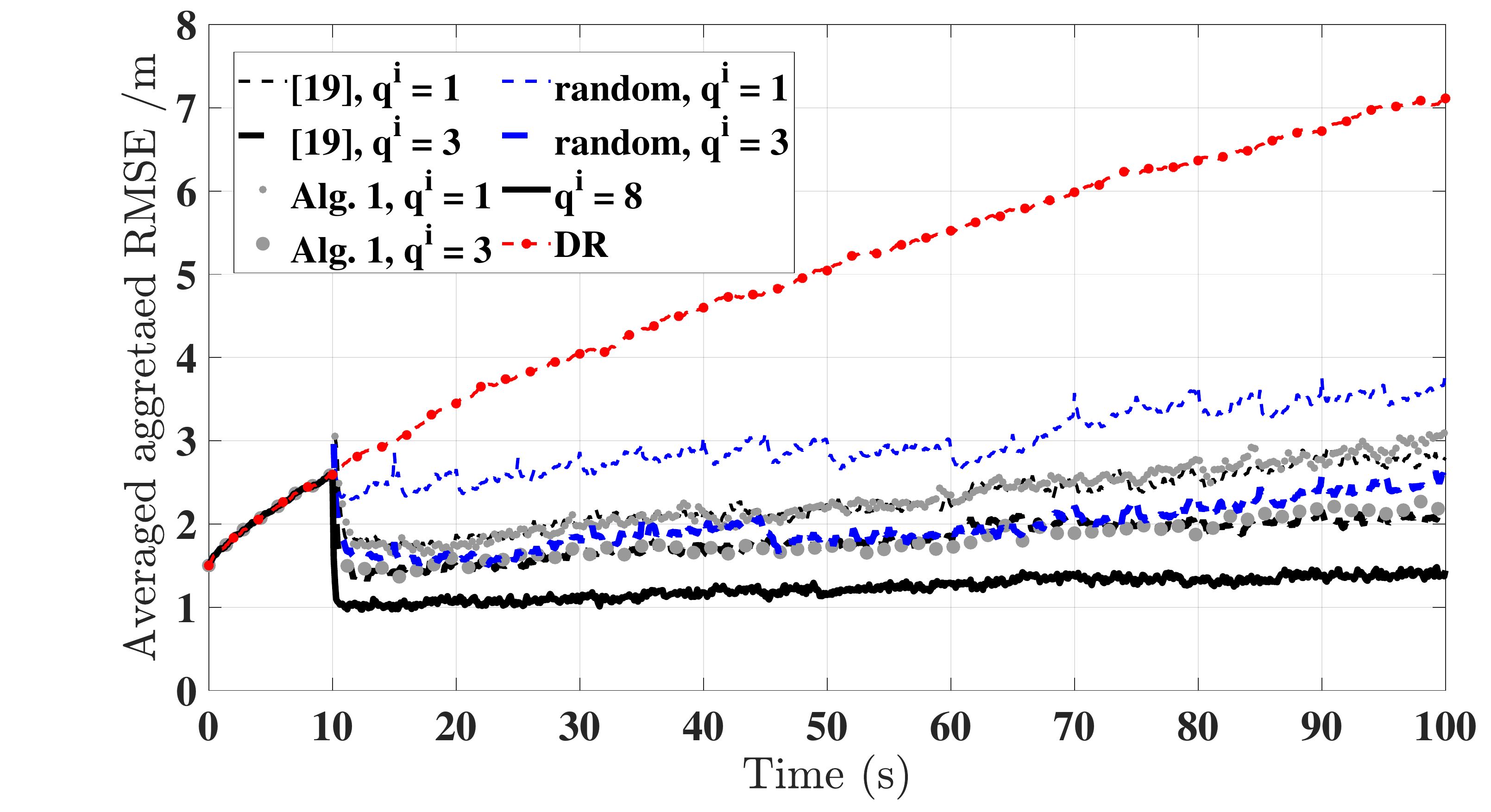}
	\vspace{-0.1 in}
	\caption{Averaged aggregated RMSE result for the Monte Carlo study.}
	\label{fig::rmse_MC}
\end{figure}

\begin{figure}
	\centering
\includegraphics[width=3in]{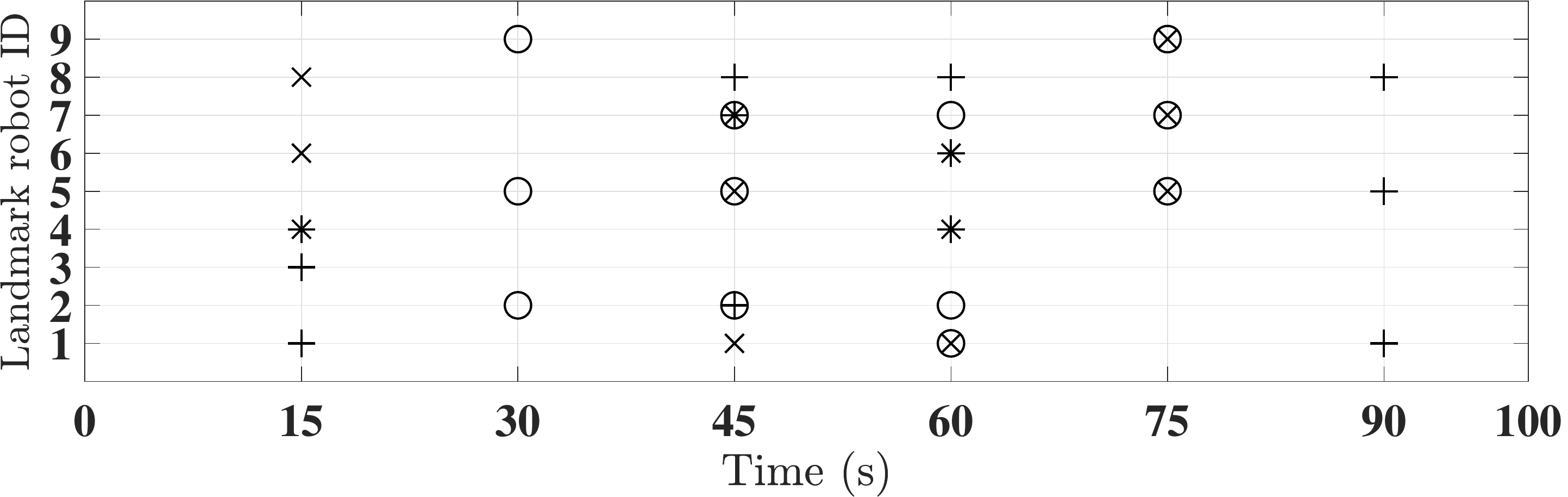}
\vspace{-0.1 in}
	\caption{Landmark selection for one of the Monte Carole simulation cases by robot $3$ (shown by $\times$) robot $6$ (shown by $\textup{o}$) and robot $9$ (shown by $+$) when they implement Algorithm~\ref{alg::subopt} to select $q^i=3$ landmark robots.
	For clarity of presentation, we have only shown the landmark selection at some selected timesteps.
}
	\label{fig::landmark_selection_MC}
\end{figure}

Table~\ref{table::CL_time} shows the relative measurement scenario we implement. This table specifies the robots that can take relative measurements at time intervals during this simulation. For this study, we assume that when a robot is allowed to take relative measurements, it can potentially take measurement with respect to all the other robots in the team. Figure~\ref{fig::logdet_MC} shows the time history of the logarithm of averaged determinant from $M = 50$ Monte Carlo simulations ($\textup{log}(\frac{1}{M}\sum_{i=1}^{M}\text{det}(\boldsymbol{P}_c^+(k)))$ while Fig.~\ref{fig::rmse_MC} shows the averaged aggregated RMSE result  ($\frac{1}{M}\sum_{i=1}^M\sum_{j=1}^N(\boldsymbol{x}^j(k)-\boldsymbol{\hat{x}^{j+}}(k))^2,j\in\mathcal{V}$).
We observe the same trend as we reported for  Fig.~\ref{fig::logdet_dataset} for the localization accuracy of the CL when we implement Algorithm~1 in contrast to when we implement the landmark selection method of~\cite{tzoumas2017scheduling} and when we process all the available relative measurements as well as when we adopt the random selection approach. 
Also Fig.~\ref{fig::rmse_MC} 
shows that CL significantly outperforms dead-reckoning (DR) only localization and the random selection, particularly in the case of $q^i = 1$ where the available resources are very limited.
Table~\ref{table::comp_time} shows the average execution time of the scheduling algorithm in \cite{tzoumas2017scheduling} and Algorithm~\ref{alg::subopt} for sample cases from our Monte-Carlo simulations corresponding to different values of $q^i$. We also included the execution time when number of robots are increased to $N=15$. As seen, Algorithm~\ref{alg::subopt} is substantially more cost effective than algorithm of~\cite{tzoumas2017scheduling} and also  has better scalability with respect to $q^i$ and $N$.
Finally, Fig.~\ref{fig::landmark_selection_MC} shows the landmark robots selection result by Algorithm~\ref{alg::subopt}  at some selected timesteps for robots $3$, $6$ and $9$ for one of our Monte Carlo simulation cases. As we can see, for each robot the algorithm chooses different landmark robots at different timesteps to yield a better localization performance (similar trend is observed for the other robots but are not shown here for brevity).


\begin{table}[h]
\centering
\caption{Average execution time of measurement selection method from a sample case of Monte Carlo simulation.}
\label{table::comp_time}
\begin{tabular}{|c|c|c|c|c|}
\hline
 &  & & \multicolumn{2}{c|}{Run time per robot (ms)} \\ \hline
CPU  & $N$ & $q^i$ &\cite{tzoumas2017scheduling} & Algorithm~\ref{alg::subopt} \\ \hline
\multirow{6}{*}{\scalebox{0.8}{Intel  Core$^{TM}$ i7-9750H@2.6GHz}}  & \multirow{3}{*}{9} & 1 & 19.42 & 3.05 \\ \cline{3-5}
&  & 3 & 52.66 & 3.09 \\ \cline{3-5} 
&  & 5 & 75.42 & 3.18 \\ \cline{2-5}
& \multirow{3}{*}{15} & 2 & 555.9 & 8.52  \\ \cline{3-5}
&  & 5 & 1209 & 8.53 \\ \cline{3-5} 
&  & 8 & 1524  & 8.57 \\ \cline{3-5}
\hline
\end{tabular}
\end{table}

\section{Conclusions}
\label{sec::con}
In this paper, we studied the problem of lowering the communication and computation costs of a decentralized CL algorithm by relative measurement scheduling. 
We provided a novel method that allows each robot to choose its restricted number of landmark robots locally without any collaboration with other team remembers. Our propose solution does not require full-observability, and has a polynomial time complexity. As a result our proposed algorithm can be a practical solution for real-time implementation for robotic teams with resource constrained~robots. 
Our future work includes extending our results to landmark selection for  loosely couple CL algorithms where the cross-covariance terms are not maintained but accounted for  implicitly.

\bibliographystyle{IEEEtran}
\bibliography{IEEEabrv,main_arxiv.bib}

\appendix
\renewcommand{\theequation}{A.\arabic{equation}}
\renewcommand{\thelemma}{A.\arabic{lemma}}

The  auxiliary lemmas below are used in development of our main result.
\begin{lemma}
	Let $\boldsymbol{A} \in \mathbb{R}^{n\times n}$ be a positive semi-definite matrix. Then, ${\rm det}(\boldsymbol{I}_n + \boldsymbol{A}) \geq 1 + {\rm tr}(\boldsymbol{A}) > 0$.
	\label{lem::det_tr_ineq_0}
\end{lemma}
\begin{IEEEproof}
Let $\{\lambda_i\}_{i=1}^n$ be the set of the eigenvalues of $\boldsymbol{A}$. Then, the eigenvalues of  $(\boldsymbol{I}_n+\boldsymbol{A})$ are $\{1+\lambda_i\}_{i=1}^n$. Thus,
	${\rm det}(\boldsymbol{I}_n + \boldsymbol{A}) = \prod_{i=1}^{n}(1+\lambda_i) = 1 + \lambda_1 + \lambda_2 + \cdots + \lambda_n + O(\lambda_i^2) = 1 + {\rm tr}(\boldsymbol{A})+ O(\lambda_i^2)$. Since  $\boldsymbol{A}$ is positive semi-definite, we have  $\lambda_i \geq 0$, $i\in\{1,\cdots,n\}$. Therefore, the $O(\lambda_i^2)$ terms in ${\rm det}(\boldsymbol{I}_n + \boldsymbol{A})$ are all non-negative. Moreover, ${\rm tr}(\boldsymbol{A}) \geq 0$. As a result, ${\rm det}(\boldsymbol{I}_n + \boldsymbol{A}) \geq 1 + {\rm tr}(\boldsymbol{A})>0$.
\end{IEEEproof}
\medskip
\begin{lemma}\label{lem::det_tr_ineq_1}
	Consider $\boldsymbol{A},\boldsymbol{B},\boldsymbol{C}\in\mathbb{R}^{n \times n}$ with $\vect{A}>\vect{0}_n$ and 
	$\boldsymbol{M} = \begin{bmatrix}
	\boldsymbol{A} & \boldsymbol{B}^\top \\ \boldsymbol{B} & \boldsymbol{C}
	\end{bmatrix} \geq \boldsymbol{0}_{2n}$.
	Then, ${\rm tr}(\boldsymbol{A} + \boldsymbol{C} - \boldsymbol{B} - \boldsymbol{B}^\top) \geq {\rm tr}(\boldsymbol{A} + \boldsymbol{B}\boldsymbol{A}^{-1}\boldsymbol{B}^\top - \boldsymbol{B} - \boldsymbol{B}^\top) \geq  0$.
\end{lemma}
\begin{IEEEproof}
	Using congruent transformation with invertible transformation matrix $\boldsymbol{T} = \begin{bmatrix}
	\boldsymbol{I}_n & \boldsymbol{I}_n \\ \boldsymbol{0}_n & -\boldsymbol{I}_n
	\end{bmatrix}$
	, we obtain
	\begin{equation*}
	\boldsymbol{\bar{M}} = \boldsymbol{T}^\top \boldsymbol{M} \boldsymbol{T} = \begin{bmatrix}
	 \boldsymbol{A} & \boldsymbol{A} - \boldsymbol{B}^\top \\
	 \boldsymbol{A} - \boldsymbol{B} & \boldsymbol{A} + \boldsymbol{C} - \boldsymbol{B} - \boldsymbol{B}^\top
	\end{bmatrix} \geq \boldsymbol{0}_{2n},
	\end{equation*}
	due to the matrix congruence property~(see \cite[Theorem 8.1.17]{golub2013matrix}).
	Then, due to the positive semidefiniteness of $\boldsymbol{\bar{M}}$, we have  
	$\boldsymbol{A} + \boldsymbol{C} - \boldsymbol{B} - \boldsymbol{B}^\top \geq \boldsymbol{0}_n$.
	Also, by Schur complement~\cite[Thoerem 14.8.4]{harville1997matrix} for $\boldsymbol{M}$, we have 
	$\boldsymbol{C} - \boldsymbol{B}\boldsymbol{A}^{-1}\boldsymbol{B}^\top \geq \boldsymbol{0}_n$, which guarantees that ${\rm tr}(\boldsymbol{C})\geq {\rm tr}(\boldsymbol{B}\boldsymbol{A}^{-1}\boldsymbol{B}^\top)$.
	Then, we can deduce that ${\rm tr}(\boldsymbol{A} + \boldsymbol{C} - \boldsymbol{B} - \boldsymbol{B}^\top) \geq {\rm tr}(\boldsymbol{A} + \boldsymbol{B}\boldsymbol{A}^{-1}\boldsymbol{B}^\top - \boldsymbol{B} - \boldsymbol{B}^\top)$.
	Next, we show that ${\rm tr}(\boldsymbol{A} + \boldsymbol{B}\boldsymbol{A}^{-1}\boldsymbol{B}^\top - \boldsymbol{B} - \boldsymbol{B}^\top)\geq0$.
	Let $\boldsymbol{\check{M}} = \begin{bmatrix}
	\boldsymbol{A} & \boldsymbol{B}^\top \\ \boldsymbol{B} & \boldsymbol{B}\boldsymbol{A}^{-1}\boldsymbol{B}^\top
	\end{bmatrix}$.
	Since $\boldsymbol{A} > \boldsymbol{0}_n$ and $\boldsymbol{B}\boldsymbol{A}^{-1}\boldsymbol{B}^\top - \boldsymbol{B}\boldsymbol{A}^{-1}\boldsymbol{B}^\top = \boldsymbol{0}_n$, by Schur complement property of $\boldsymbol{\check{M}}$ 
	we obtain $\boldsymbol{\check{M}} \geq \boldsymbol{0}_{2n}$.
	Following similar congruent transformation procedures above, we can prove $\boldsymbol{A} + \boldsymbol{B}\boldsymbol{A}^{-1}\boldsymbol{B}^\top - \boldsymbol{B} - \boldsymbol{B}^\top \geq \boldsymbol{0}_n$, which guarantees that ${\rm tr}(\boldsymbol{A} + \boldsymbol{B}\boldsymbol{A}^{-1}\boldsymbol{B}^\top - \boldsymbol{B} - \boldsymbol{B}^\top) \geq 0$, completing the proof.
\end{IEEEproof}

\end{document}